\documentclass[12pt,journal,compsoc,onecolumn]{IEEEtran}

 \usepackage[T1]{fontenc}

\usepackage{lmodern}
\usepackage{graphicx}
\usepackage{microtype}
\usepackage{subfigure}
\usepackage{booktabs,pifont,multirow,multicol} 
\usepackage{amsmath}
\usepackage{amssymb}
\usepackage{mathtools}
\usepackage{pifont}
\usepackage[noadjust]{cite}
\usepackage{caption}

\usepackage{multirow}
\newcommand{\cmark}{\ding{51}}
\newcommand{\xmark}{\ding{55}}

\usepackage[textsize=tiny]{todonotes}

\def\x{{\mathbf x}}

\def\UU{{\bf U}}

\def\x{{\mathbf x}}

\def\U{{\mathbf U}} 
\def\V{{\mathbf V}} 
\def\A{{\mathbf \Lambda}}

\def\A{{\bf A}}

\def\U{{\bf U}}
\def\W{{\bf W}}
\def\WW{{\bf W}}
\def\SS{{\cal S}}

\def\N{{\cal N}}
\def\G{{\cal G}}
\def\V{{\cal V}}
\def\E{{\cal E}}
\def\F{{\cal F}}

\def\M{{\bf M}}

\def\P{{\bf P}}

\usepackage[ruled,vlined]{algorithm2e}

\title{Designing Semi-Structured Pruning of Graph Convolutional Networks for Skeleton-based Recognition}

\author{Hichem Sahbi
\vspace{1cm}
\\ 
{Sorbonne University, CNRS, LIP6, F-75005, Paris, France}}

\begin{document}
 \maketitle
\begin{abstract}

Deep neural networks (DNNs) are nowadays witnessing a major success in solving many pattern recognition tasks including skeleton-based classification. The deployment of DNNs on edge-devices, endowed with limited time and memory resources, requires designing lightweight and efficient variants of these networks. Pruning is one of the lightweight network design techniques that operate by removing unnecessary network parts, in a structured or an unstructured manner, including individual weights, neurons or even entire channels. Nonetheless, structured and unstructured pruning methods, when applied separately, may either be inefficient or ineffective.\\
In this paper,  we devise a novel semi-structured method that discards the downsides of structured and unstructured pruning while gathering their upsides to some extent. The proposed solution is based on a differentiable cascaded parametrization which combines (i) a band-stop mechanism that prunes weights depending on their magnitudes, (ii) a weight-sharing parametrization that prunes connections  either individually or group-wise, and (iii) a gating mechanism which arbitrates between different group-wise and entry-wise pruning. All these cascaded parametrizations are built upon a common latent tensor which is trained end-to-end  by minimizing a classification loss and a surrogate tensor rank regularizer. Extensive experiments, conducted on the challenging tasks of action and hand-gesture recognition, show the clear advantage of our proposed semi-structured pruning approach against both structured and unstructured pruning, when taken separately, as well as the related work. 

\noindent {\bf keywords.} {Structured and unstructured pruning  \and Semi-structured pruning \and Graph-convolutional networks \and Skeleton-based recognition.}
\end{abstract}

\section{Introduction}
 Deep  neural networks (DNNs) are nowadays  becoming a hotspot in machine learning with increasingly performant models used to  approach eclectic  pattern recognition tasks~\cite{Krizhevsky2012,jiu2017nonlinear,jiu2019deep}.  These  models are also steadily  oversized and this makes their deployment on cheap devices,  endowed with limited hardware resources,  very challenging.  In particular, hand-gesture recognition and human computer interaction tasks require fast and lightweight DNNs with high recognition performances.   However, DNNs are currently showing  some saturated improvement in accuracy while their computational efficiency remains a major issue.  Among these DNN models, graph convolutional networks (GCNs) are deemed effective  especially  on non-euclidean domains including  skeleton-data~\cite{Zhua2016}.  Two families of GCNs exist in the literature:  spectral and spatial.  Spectral methods  project graph signals from the input to the  Fourier domain,  achieve convolution,  prior  to back-project the convolved signals  in the input domain \cite{kipf17,Li2018,sahbiiccv21,mazari2019mlgcn}.   Spatial methods proceed differently  by aggregating signals through neighboring nodes, using multi-head attention,  prior to achieve convolutions (as inner products) on the resulting node aggregates \cite{Gori2005,attention2019,sahbi2010context,sahbi2021learning,sahbi2021kernel}.  Spatial GCNs are known to be more effective compared to spectral ones.  Nonetheless,  with multi-head attention,  spatial GCNs  become oversized, computationally  overwhelming,  and their deployment of cheap devices requires  making them  lightweight and still effective \cite{sahbi2021lightweight,sahbi2023phase}.  \\

\indent Several  existing works address the issue of  lightweight network design, including tensor decomposition~\cite{howard2019},  quantization~\cite{DBLP:journals/corr/HanMD15},  distillation~\cite{DBLP:conf/aaai/MirzadehFLLMG20,sahbi2006hierarchy},  neural architecture search \cite{nasprun} and pruning~\cite{DBLP:conf/nips/CunDS89,DBLP:conf/nips/HassibiS92,DBLP:conf/nips/HanPTD15,sahbi2022topologically}.  Pruning methods are particularly  effective,  and their general recipe consists in removing  connections  in order to enable reduced storage and faster inference with a minimal impact on classification performances. One of the mainstream methods is magnitude pruning (MP) \cite{DBLP:journals/corr/HanMD15} which removes the smallest weight connections before retraining the pruned networks. Two categories of MP techniques  exist in the literature: unstructured ~\cite{DBLP:conf/nips/HanPTD15,DBLP:journals/corr/HanMD15}  and structured~\cite{DBLP:conf/iclr/0022KDSG17,DBLP:conf/iccv/LiuLSHYZ17}.   Unstructured methods remove weights individually by ranking them according to the importance of their magnitudes whilst structured approaches zero-out groups of weights (belonging to entire rows, columns, filters or channels) according to the importance of their {\it aggregated} magnitudes.  Unstructured MP results into more flexible, accurate networks, and allows reaching any fine-grained targeted pruning rate but requires dedicated hardware to actually achieve efficient computation. In contrast, structured MP offers a more practical advantage by making the resulting DNNs compatible with standard hardware for efficient computation. However, this comes at the expense of a reduced classification performance and coarse-grained pruning rates. {\it In order to fully exhibit the potential of these two pruning categories, a more suitable framework should gather the upsides of both structured and unstructured pruning while discarding their downsides to some extent.}   \\

 \indent In this paper,  we  introduce a novel variational MP approach that leverages both structured and unstructured pruning.  This method dubbed as {\it semi-structured} is based on a differentiable  cascaded weight parametrization composed of (i) a band-stop mechanism enforcing the prior that the smallest weights should be removed,  (ii) a weight-sharing that groups mask entries belonging to the same rows,  columns, or channels in a given tensor, and (iii) a gating mechanism which arbitrates between different mask group assignments while maximizing the accuracy of the trained  lightweight networks.  We also consider a budget loss that allows implementing any targeted fine-grained pruning rate and  reducing further the rank of the pruned tensors, resulting into more efficient networks while being closely accurate as shown later in experiments.

 \section{Related work}

The following review discusses the related work in variational pruning and skeleton-based recognition, highlighting the limitations that motivate our contributions.\\

\noindent {\bf Variational Pruning.} The general concept behind variational pruning is to learn weights and binary masks that capture the topology of pruned networks. This is achieved by minimizing a global loss that combines a classification error and a regularizer that controls the sparsity (or the cost) of the resulting networks \cite{DBLP:conf/iccv/LiuLSHYZ17,REFWen,REFICLR}.   However, these approaches are powerless to implement any given targeted pruning rate without overtrying multiple weighting of the regularizers. Alternative methods explicitly model the network cost using $\ell_0$-based criteria \cite{REFICLR,REFDrop}  in order to minimize the discrepancy between the observed and the targeted costs.  Existing solutions rely on sampling heuristics or relaxation, which promote sparsity --- using different regularizers ($\ell_1$/$\ell_2$-based,  entropy,  etc. )  \cite{REFGordon,REFCarreira,refref74,refref75} --- but are powerless to implement any given targeted cost exactly and result in overpruning effects leading to disconnected subnetworks. Furthermore, most of the existing solutions, including magnitude pruning \cite{DBLP:journals/corr/HanMD15},   decouple the training of network topology from weights, making the learning of pruned networks suboptimal.  On another hand,   existing pruning methods are either structured~\cite{DBLP:conf/iclr/0022KDSG17,DBLP:conf/iccv/LiuLSHYZ17} or unstructured~\cite{DBLP:conf/nips/HanPTD15,DBLP:journals/corr/HanMD15} so their benefit is not fully explored. In contrast to the aforementioned related work, our contribution in this paper seeks to leverage the advantage of both structured and unstructured pruning where the training of masks and weights are coupled on top of shared latent parameters.\\

\noindent {\bf Skeleton-based recognition.}   This task has gained increasing interest due to the emergence of sensors like Intel RealSense   and Microsoft Kinect.  Early methods for hand-gesture and action recognition used RGB~\cite{wang2014bags,wang2013directed,refref18,yuan2012mid}, depth~\cite{refref39},  shape / normals~\cite{refref40,refref41,Yun2012,Ji2014,Li2015a,refref59,sahbi2007kernel,sahbi2004kernel},  and skeleton-based techniques \cite{Wang2018c}. These methods were based on modeling human motions using handcrafted features \cite{Yang2014}, dynamic time warping~\cite{Vemulapalli2014},  temporal information \cite{refref61,refref11}, and temporal pyramids~\cite{Zhua2016}. However, with the resurgence of deep learning, these methods have been quickly overtaken by 2D/3D Convolutional Neural Networks (CNNs)~\cite{refref10,REF3,mazari2019deep}, Recurrent Neural Networks (RNNs) \cite{Zhua2016,Du2015,Liu2016,DeepGRU,Zhang2017,GCALSTM},  manifold learning~\cite{Huangcc2017,ref23,Liu2021,RiemannianManifoldTraject},  attention-based networks \cite{Song2017},  and GCNs \cite{Lib2018,Yanc2018,Wen2019,Jiang2020}.  The recent emergence of GCNs, in particular,  has led to their increased use in skeleton-based recognition \cite{Li2018}. These models capture spatial and temporal attention among skeleton joints with better interpretability.  However, when tasks involve relatively large input graphs, GCNs (particularly with multi-head attention) become computationally inefficient and require lightweight design techniques.  In this paper, we design efficient GCNs that make skeleton-based recognition highly efficient while also being effective.

\section{A Glimpse on Graph Convolutional Networks}

Let $\SS=\{\G_i=(\V_i, \E_i)\}_i$ denote a collection of graphs with $\V_i$, $\E_i$ being respectively the nodes and the edges of $\G_i$. Each graph $\G_i$ (denoted for short as $\G=(\V, \E)$) is endowed with a signal $\{\phi(u) \in \mathbb{R}^s: \ u \in \V\}$ and associated with an adjacency matrix $\A$. GCNs aim at learning a set of $C$ filters $\F$ that define convolution on $n$ nodes of $\G$ (with $n=|\V|$) as $(\G \star \F)_\V = f\big(\A \  \UU^\top  \   \W\big)$, here $^\top$ stands for transpose,  $\UU \in \mathbb{R}^{s\times n}$  is the  graph signal, $\W \in \mathbb{R}^{s \times C}$  is the matrix of convolutional parameters corresponding to the $C$ filters and  $f(.)$ is a nonlinear activation applied entry-wise. In  $(\G \star \F)_\V$, the input signal $\UU$ is projected using $\A$ and this provides for each node $u$, the  aggregate set of its neighbors. Entries of $\A$ could be handcrafted or learned so  $(\G \star \F)_\V$  corresponds to  a convolutional block with two layers; the first one aggregates signals in $\N(\V)$ (sets of node neighbors) by multiplying $\UU$ with $\A$ while the second layer achieves convolution by multiplying the resulting aggregates with the $C$ filters in $\W$. Learning  multiple adjacency (also referred to as attention) matrices (denoted as $\{\A^k\}_{k=1}^K$) allows us to capture different contexts and graph topologies when achieving aggregation and convolution.  With multiple matrices $\{\A^k\}_k$ (and associated convolutional filter parameters $\{\W^k\}_k$),  $(\G \star \F)_\V$ is updated as $f\big(\sum_{k=1}^K \A^k   \UU^\top     \W^k\big)$. Stacking aggregation and convolutional layers, with multiple  matrices $\{\A^k\}_k$, makes GCNs accurate but heavy. We propose, in what follows, a method that makes our networks lightweight and still effective.

\section{Proposed Method: Semi-Structured Magnitude Pruning}
In what follows, we formally subsume a given GCN as a multi-layered neural network $g_\theta$  whose weights are defined as $\theta =  \left\{\WW^1,\dots, \WW^L \right\}$, being $L$ its depth,  $\WW^\ell \in \mathbb{R}^{d_{\ell-1} \times d_{\ell}}$ its $\ell^\textrm{th}$  layer weight tensor, and $d_\ell$ the dimension of $\ell$. The output of a given layer  $\ell$ is defined as
$ \mathbf{\phi}^{\ell} = f_\ell({\WW^\ell}^\top \  \mathbf{\phi}^{\ell-1})$, $\ell \in \{2,\dots,L\}$,  with $f_\ell$ an activation function; without a loss of generality, we omit the bias in the definition of  $\mathbf{\phi}^{\ell}$.\\
 \noindent Pruning consists in zeroing-out a subset of weights in $\theta$ by multiplying $\WW^\ell$ with a binary mask $\M^\ell \in \{ 0,1 \}^{d_{\ell-1} \times d_{\ell}}$.
The binary entries of  $\M^\ell$ are set depending on whether the underlying layer connections are pruned, so $\mathbf{\phi}^{\ell} = f_\ell( (\M^\ell \odot \WW^\ell )^\top \ \mathbf{\phi}^{\ell-1} )$, here $\odot$ stands for the element-wise matrix product.  In our definition of  semi-structured pruning, entries of the tensor $\{\M^\ell\}_\ell$ are set depending on the prominence and also on how the underlying connections in $g_\theta$ are grouped; pruning that removes the entire connections individually (resp. jointly) is referred to as {\it unstructured} (resp. {\it structured}) whereas pruning that removes some connections independently and others jointly is dubbed as {\it semi-structured.}  However, such pruning (with $\{\M^\ell\}_\ell$) suffers from several drawbacks. In the one hand, optimizing the discrete set of variables $\{\M^\ell\}_\ell$ is deemed highly combinatorial and intractable especially on large networks. In the other hand,  the total number of parameters $\{\M^\ell\}_\ell$, $\{\WW^\ell\}_\ell$ is twice the number of connections in $g_\theta$ and this increases training complexity and may also lead to overfitting.

\subsection{Semi-Structured Weight Parametrization} In order to overcome the aforementioned issues, we consider an alternative {\it parametrization} that allows finding both the topology of the pruned networks together with their weights, without doubling the size of the training parameters, while making magnitude pruning semi-structured and learning still effective. This parametrization corresponds to the Hadamard product involving a weight tensor and a {\it cascaded}  function applied to the same tensor as
\begin{eqnarray}\label{eq2} 
  \WW^\ell = \hat{\WW}^\ell \odot \big[\psi_3 \circ \psi_2 \circ \psi_1 (\hat{\WW}^\ell)\big],
\end{eqnarray}
\noindent being $\hat{\WW}^\ell$ a latent tensor and $\psi(\hat{\WW}^\ell)$ (with $\psi=\psi_3 \circ \psi_2 \circ \psi_1$) a continuous relaxation of $\M^\ell$ which enforces the prior that (i) smallest weights $\hat{\WW}^\ell$ should be removed from the network, (ii) the underlying mask entries $\psi(\hat{\WW}^\ell)$ are shared (across tensor rows, columns, channels, etc.) when pruning is structured,  and (iii) any given mask entry  in  $\psi(\hat{\WW}^\ell)$ is either unstructurally or structurally  pruned.  In what follows,  we detail the different parametrizations used to define $\psi(\hat{\WW}^\ell)$; unless explicitly mentioned, we omit $\ell$ in the definition of  $\hat{\WW}^\ell$ and we rewrite it simply as $\hat{\WW}$.\\  

\noindent {\bf Band-stop Parametrization ($\psi_1$).} This parametrization $\psi_1$ is entry-wise applied to the tensor  $\hat{\WW}$ and enforces the prior that smallest weights should be removed from the network. In order to achieve this goal, $\psi_1$ must be (i) bounded in $[0,1]$, (ii) differentiable, (iii) symmetric, and (iv) $\psi_1(\omega) \leadsto 1$ when $|\omega|$ is sufficiently large and $\psi_1(\omega) \leadsto 0$ otherwise. The first and the fourth properties ensure that the parametrization is neither acting as a scaling factor greater than one nor changing the sign of the latent weight, and also acts as the identity for sufficiently large weights, and as a contraction factor for small ones. The second property is necessary to ensure that $\psi_1$ has computable gradient while the third condition guarantees that only the magnitudes of the latent weights matter. A  choice, used in practice, that satisfies these four conditions is
\begin{equation} \label{eq0} 
  \psi_1(\omega)=2 \big(1+ \exp (-\sigma \omega^2)\big)^{-1}-1,
 \end{equation}  
being $\sigma$  a scaling factor that controls the crispness (binarization) of mask entries in $\psi_1 (\hat{\WW})$.   According to Eq.~\ref{eq0},  $\sigma$ controls the smoothness of $\psi_1$ around the support of the latent weights. This allows implementing an annealed (soft) thresholding function that cuts-off all the connections in smooth and differentiable manner as training of the latent parameters evolves.  The asymptotic behavior of $\psi_1$   --- that allows selecting the topology of the pruned subnetworks  --- is obtained as training reaches the latest epochs, and this makes mask entries, in $\psi_1 (\hat{\WW})$, crisp and (almost) binary. This mask  $\psi_1 (\hat{\WW})$ (rewritten for short as $\psi_1$) is used as input to the subsequent parameterizations $\psi_2$ and $\psi_3$ as shown below. \\

\begin{figure}[hpbt]
   \begin{center}
     \centerline{\scalebox{0.81}{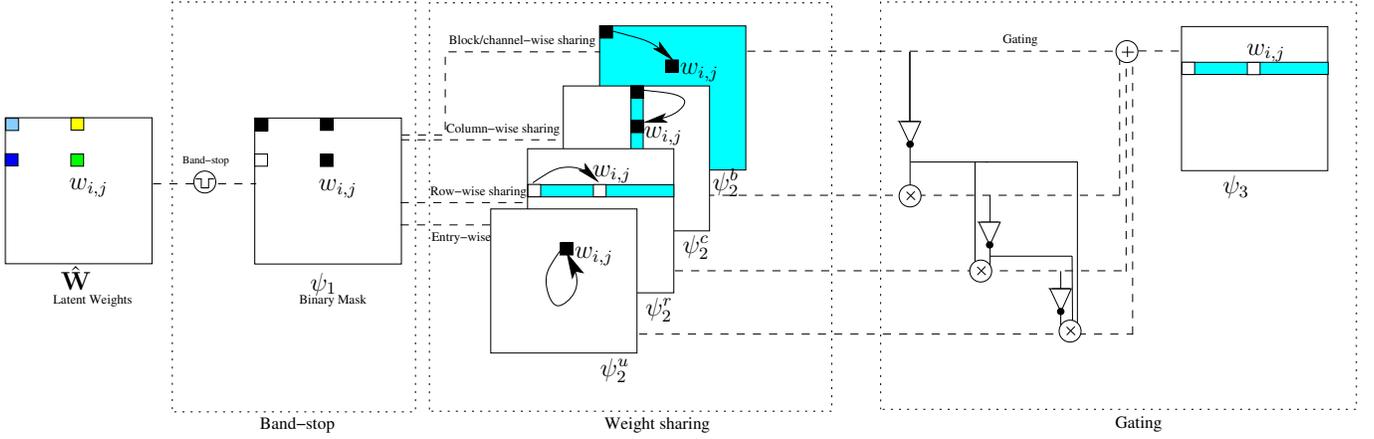}}
     \caption{This figure shows the three stages of the cascaded parametrization including (i) band-stop, (ii) weight-sharing and (iii) gating. Cyan stands for shared connections, and the triangle for the ``not gate'' operator. For ease  of visualization, only 4 connections are shown during the whole evaluation of the parameterization, and only the outcome (1 or 0) of $w_{i,j}$ is shown in the final mask tensor.} \label{fig1}
 \end{center}
 \end{figure}

\noindent {\bf Weight-sharing Parametrization ($\psi_2$).} This parametrization  $\psi_2$ implements semi-structured pruning by {\it tying} mask entries belonging to the same rows, columns or channels in the tensor $\psi_1$. More precisely, each mask entry in $\psi_2(\psi_1)$ will either be  (i) entry-wise evaluated (dependent only on its underlying weight), or (ii) shared through multiple latent weights belonging to the same row, column or channel of $\psi_1$ resulting into the following multi-head parametrization (see Fig.~\ref{fig1})
\begin{equation} 
\psi_2(\psi_1) =\begin{cases}\psi_2^u(\psi_1) \ = \psi_1  & \ \ \ \textrm{unstructured (entry-wise)} \\
\psi_2^r(\psi_1) \ = {\bf vec}^{-1}(\P_r \  {\bf vec}(\psi_1))   & \ \ \ \textrm{structured (row-wise)}   \\
\psi_2^c(\psi_1) \ = {\bf vec}^{-1}({{\bf vec} (\psi_1)^\top \ \P_c})    &\ \ \ \textrm{structured (column-wise)} \\
\psi_2^{b}(\psi_1) \ = {\bf vec}^{-1}(\P_{r}\P_c^\top \ {\bf vec} (\psi_1))  & \ \ \  \textrm{structured (block/channel-wise),} 
\end{cases}
\end{equation}
here ${\bf vec}$ (resp. ${\bf vec}^{-1}$)  reshapes a matrix into a vector (resp. vice-versa), and  $\P_r \in \{0,1\}^{(d_{\ell-1}\times d_{\ell})^2}$, $\P_c \in \{0,1\}^{(d_{\ell-1} \times d_\ell)^2}$ are two adjacency matrices that model the neighborhood system across respectively  the rows and the columns of $\psi_1$  whilst $\P_{r} \P^\top_{c}\in \{0,1\}^{(d_{\ell-1} \times d_\ell)^2}$ models this neighborhood through blocks/channels of  $\psi_1$.  When composed (with $\psi_1$), the mask $\psi_2$ inherits all the aforementioned fourth properties: mask entries in $\psi_2(\psi_1)$ remain bounded in $[0,1]$, differentiable, symmetric, and close to $1$ when entries of the latent tensor $\hat{\WW}$ (i.e., inputs of $\psi_1$) are sufficiently large and $0$ otherwise. \\    

\noindent {\bf Gating Parametrization ($\psi_3$).} As each connection in $g_\theta$ is endowed with a multi-head parametrization $\psi_2$,   we define $\psi_3$ as a gating mechanism that selects only one of them. More precisely, each mask entry can either be (i) entry-wise pruned, i.e., {untied}, or (ii) {tied} to its row, column or block/channel. Again with $\psi_3$, the composed parametrization $\psi_3 (\psi_2)$ is bounded in $[0,1]$, differentiable, symmetric and reaches  $1$ if the entries of the latent tensor $\hat{\WW}$ are sufficiently large, and $0$ otherwise. Formally, the gating mechanism $\psi_3$  is defined as   
\begin{equation}\label{eq5555}
 \psi_3(\psi_2) = \underbrace{\psi_2^{b}}_{\textrm{block-wise}} \ + \   \underbrace{(\bar{\psi}_2^{b}) \odot \psi_2^{c}}_{\textrm{column-wise}}\ + \ \underbrace{(\bar{\psi}_2^{b} \odot  \bar{\psi}_2^{c}) \odot  \psi_2^{r}}_{\textrm{row-wise}}  \ + \  \underbrace{(\bar{\psi}_2^{b} \odot  \bar{\psi}_2^{c} \odot  \bar{\psi}_2^{r}) \odot \psi_2^{u}}_{\textrm{entry-wise}},
\end{equation} 

\noindent being $\bar{\psi}_2^{b}=\U-\psi_2^b$, and $\U$ a tensor of ones with the same dimensions as $\psi_2^b$ (and  $\bar{\psi}_2^{r}$,  $\bar{\psi}_2^{c}$,  $\bar{\psi}_2^{u}$ are similarly defined). It is easy to see that when entries in $\psi_1$ (and hence $\psi_2$) are crisp, at most one of these four terms is activated (i.e., equal to one) for each connection in $g_\theta$.  From Eq.~\ref{eq5555}, block-wise pruning has the highest priority, followed by column-wise, row-wise and then entry-wise pruning respectively. This priority allows designing highly efficient lightweight networks with a coarse-granularity budget implementation for block/column/row-wise (structured) pruning, while entry-wise (unstructured) pruning is less computationally efficient but allows reaching any targeted budget with a finer granularity, and thereby with a better accuracy. Note that this parametrization acts a weight regularizer which not only improves the lightweightness of the pruned networks but also their generalization performances (as shown  later in  experiments). Note also that $\psi_1$ and $\psi_2$ are commutable in the cascaded parameterization $\psi=\psi_3 \circ \psi_2 \circ \psi_1$  but $\psi_3$ should be applied at the end of the cascade.

\subsection{Budget-Aware Variational Pruning}
By considering Eq.~\ref{eq2}, we define our semi-structured pruning loss as
\begin{equation}\label{eq3}
\begin{array}{lll}
{\cal L}_e\big( \{ \psi_3\circ \psi_2 \circ \psi_1(\hat{\WW}^{\ell}) \odot \hat{\WW }^{\ell} \}_\ell\big) + \lambda \displaystyle \bigg(\sum_{\ell=1}^{L-1}  {1}_{d_\ell}^\top  [\psi_3\circ \psi_2 \circ \psi_1(\hat{\WW}^{\ell})] {1}_{d_{\ell+1}} \ - c\bigg)^2,
\end{array}     
\end{equation} 
being $1_{d_\ell}$ a vector of $d_\ell$ ones and  the left-hand side term is the cross entropy loss that measures the discrepancy between predicted and ground-truth labels. The right-hand side term is a budget loss that allows reaching any targeted pruning cost $c$. Nonetheless, it's worth noticing that actual efficiency is not only related to the pruning rate but also to the actual dimensionality of the tensors. In order to take full advantage of the semi-structured setting of our method, we complement the aforementioned budget function with another one that minimizes the rank of the pruned tensors $\{\psi_3\circ \psi_2 \circ \psi_1(\hat{\WW}^{\ell})\}_\ell$. However, as the rank is not differentiable, we consider a surrogate function (as an upper bound) of the rank. Hence, Eq.~\ref{eq3} becomes
\begin{equation}\label{eq4}
\begin{array}{lll}
  {\cal L}_e\big( \psi_3\circ \psi_2 \circ \psi_1(\hat{\WW}^{\ell}) \odot \{\hat{\WW }^{\ell} \}_\ell)& + \lambda \displaystyle \bigg(\sum_{\ell=1}^{L-1}  {1}_{d_\ell}^\top [\psi_3\circ \psi_2 \circ \psi_1(\hat{\WW}^{\ell})] {1}_{d_{\ell+1}} \ - c\bigg)^2 \\
 & +   \beta \displaystyle \sum_{\ell=1}^{L-1}  r[(\psi_3\circ \psi_2 \circ \psi_1(\hat{\WW}^{\ell})],
\end{array}     
\end{equation}
here  $r[\WW]$ is a surrogate differentiable rank function set in practice to
\begin{equation}\label{eqqqqq}
r[\WW] =  \big[1_{d_{\ell+1}}^\top-\exp(-\gamma 1_{d_\ell}^\top \WW) \big] 1_{d_{\ell+1}} + 1^\top_{d_{\ell}}       \big[1_{d_\ell} - \exp(-\gamma \WW 1_{d_{\ell+1}}) \big],
\end{equation}   
\noindent being $\gamma$ an annealed temperature and $\exp(.)$ is entry-wise applied.  Eq.~\ref{eqqqqq} seeks to minimize  the number of non-null rows/colums in a given tensor $\WW$, and this allows achieving  higher speedup compared to when {\it only} the budget loss is minimized (see experiments). In Eq.~\ref{eq4}, $\beta$ controls the ``structureness'' of pruning; large $\beta$ favors stringent tensors first through blocks, columns and then through rows, while smaller $\beta$ leads to  {\it mixed} structured and unstructured pruning. Once the above loss optimized, actual rank minimization requires reordering dimensions layer-wise in order to fully benefit from compact tensors and eliminate fragmentation; this is achievable as only outward connections,  from unpruned neurons in each layer,  are actually pruned during optimization.

\begin{table}
\begin{center} 
\resizebox{\textwidth}{!}{
  \begin{tabular}{l||l||l||l}
 \multicolumn{1}{c||}{Entry-wise}  &   \multicolumn{1}{c||}{Row-wise} &   \multicolumn{1}{c||}{Column-wise} &   \multicolumn{1}{c}{Block-wise}   \\
    \hline
    \hline
 $[{\bf J}_1]_{ij,pq}= 1_{\{ij=pq\}}  \psi'_1(\hat{\WW}_{pq})$         &   \multicolumn{1}{c||}{NA}    &  \multicolumn{1}{c||}{NA}          &   \multicolumn{1}{c}{NA}         \\
     
 $[{\bf J}^u_2]_{ij,pq}=1_{\{ij=pq\}}$         &  $[{\bf J}^r_2]_{ij,pq}=[\P_r]_{ij,pq}$        &  $[{\bf J}^c_2]_{ij,pq}=[\P_c']_{ij,pq}$            & $[{\bf J}^b_2]_{ij,pq}=[\P_r\P_c']_{ij,pq}$             \\

    $[{\bf J}^u_3]_{ij,pq}=1_{\{ij=pq\}}$ &  $[{\bf J}^r_3]_{ij,pq}=1_{\{ij=pq\}}$  &   $[{\bf J}^c_3]_{ij,pq}= 1_{\{ij=pq\}}$ &   $[{\bf J}^b_3]_{ij,pq}=1_{\{ij=pq\}} $ \\
    $\ \ \ \ \ \ \ \ \ \ \ \times [\bar{\psi}_2^{b} \odot \bar{\psi}_2^{c} \odot \bar{\psi}_2^{r}]_{pq}$ &  $  \ \ \ \ \ \ \ \ \ \ \ \times  [\bar{\psi}_2^{b} \odot  \bar{\psi}_2^{c} \odot \bar{\psi}_2^{u}]_{pq}$  &   $ \ \ \ \ \ \ \ \ \ \ \  \times [\bar{\psi}_2^{b} \odot \bar{\psi}_2^{r} \odot \bar{\psi}_2^{u}]_{pq}$ &   $ \ \ \ \ \ \ \ \ \ \ \ \times [\bar{\psi}_2^{c} \odot \bar{\psi}_2^{r} \odot \bar{\psi}_2^{u}]_{pq}$  
  \end{tabular}}
\end{center}
\caption{Jacobians of different parametrizations w.r.t. different settings; here $[{\bf J}_1]_{ij,pq}=[{\partial \psi_1}\slash{\partial \hat{\WW}}]_{ij,pq}$, $[{\bf J}^\x_2]_{ij,pq}=[{\partial \psi_2^\x}\slash{\partial \psi_1}]_{ij,pq}$ and $[{\bf J}^\x_3]_{ij,pq}=[{\partial \psi_3}\slash{\partial \psi_2^\x}]_{ij,pq}$ with  $\x \in \{u,r,c,b\}$; here $u$, $r$, $c$ and $b$ stand for entry-wise, row-wise, column-wise and block-wise respectively. It is easy to see that all these Jacobians are extremely  sparse and their evaluation is highly efficient. In this table, NA stands for not applicable as the parametrization $\psi_1$ is necessarily entry-wise.}\label{tab1} 
\end{table} 
\subsection{Optimization}
Let $\cal L$ denote the global loss in Eq.~\ref{eq4}, the update of $\{\hat{\WW}^\ell\}_\ell$ is achieved using the gradient of $\cal L$ obtained by propagating the gradients through  $g_\theta$. More precisely, considering the parametrization in Eq.~\ref{eq2}, the gradient of the global loss w.r.t. $\hat{\WW}^\ell$ is obtained as
\begin{equation} 
\frac{\partial {\cal L}}{\partial \hat{\WW}^\ell} = \frac{\partial {\cal L}}{\partial \psi(\hat{\WW}^\ell)} \ \ \frac{\partial \psi(\hat{\WW}^\ell)}{\partial \psi_2\circ \psi_1(\hat{\WW}^\ell)} \ \  \frac{\partial \psi_2\circ \psi_1(\hat{\WW}^\ell)}{\partial \psi_1(\hat{\WW}^\ell)} \ \  \frac{\partial \psi_1(\hat{\WW}^\ell)}{\partial \hat{\WW}^\ell},
\end{equation} 
\noindent here the original gradient ${\partial {\cal L}}\slash{\partial \psi(\hat{\WW}^\ell)}$ is obtained from layer-wise backpropagation, and $\frac{\partial {\cal L}}{\partial \hat{\WW}^\ell}$ is obtained by multiplying the original gradient by the three rightmost Jacobians (whose matrix forms are shown in Table~\ref{tab1}).\\ 
\noindent In the above objective function (Eq.~\ref{eq4}), $\beta=0.1$ and $\lambda$ is overestimated (to 1000 in practice) in order to make Eq.~\ref{eq4} focusing on the implementation of the budget.  As training reaches its final epochs, the budget loss  reaches its minimum and the gradient of the global objective function will be dominated by the gradient of ${\cal L}_e$ (and to some extent by the gradient of the surrogate rank function); this allows improving both classification performances and efficiency as shown subsequently.

\begin{figure*}
  \centering
\includegraphics[width=0.3\linewidth]{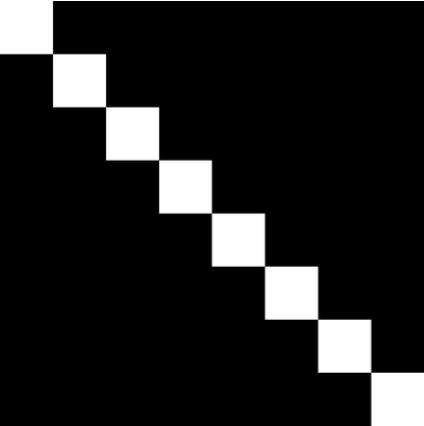}\hspace{1cm}\includegraphics[width=0.3\linewidth]{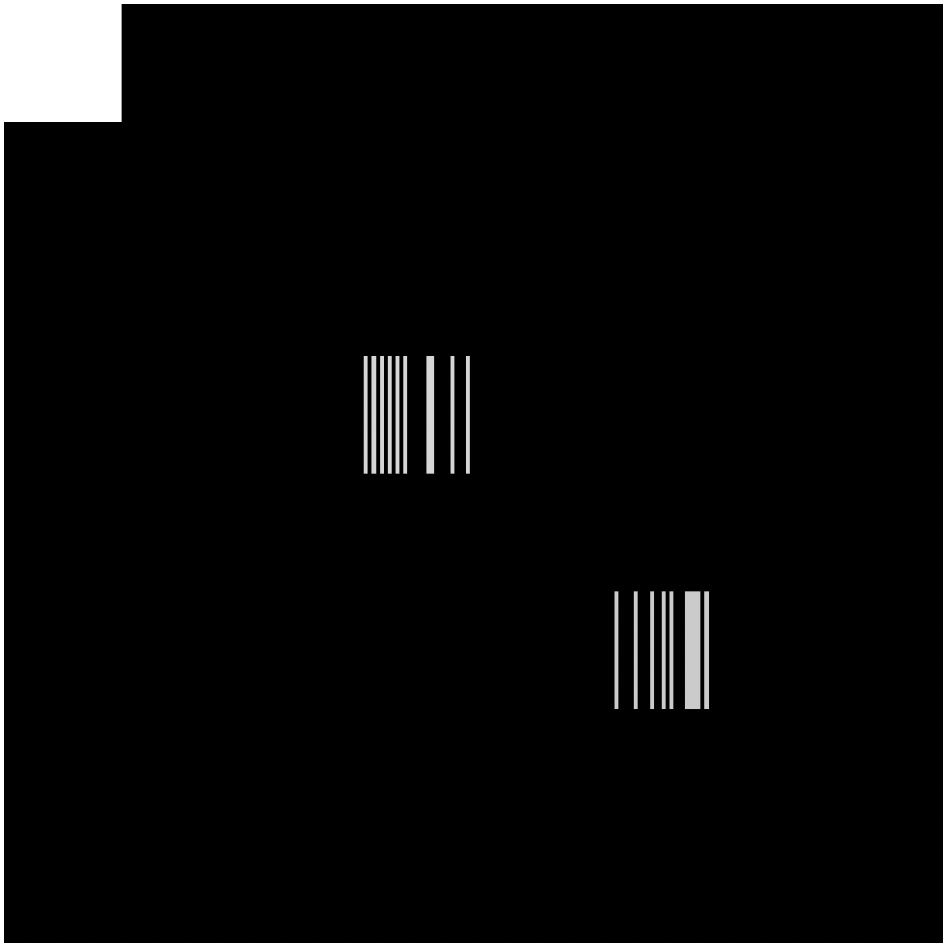}\\
(original unpruned mask) \hspace{1.5cm} (structured pruning) \\
\includegraphics[width=0.3\linewidth]{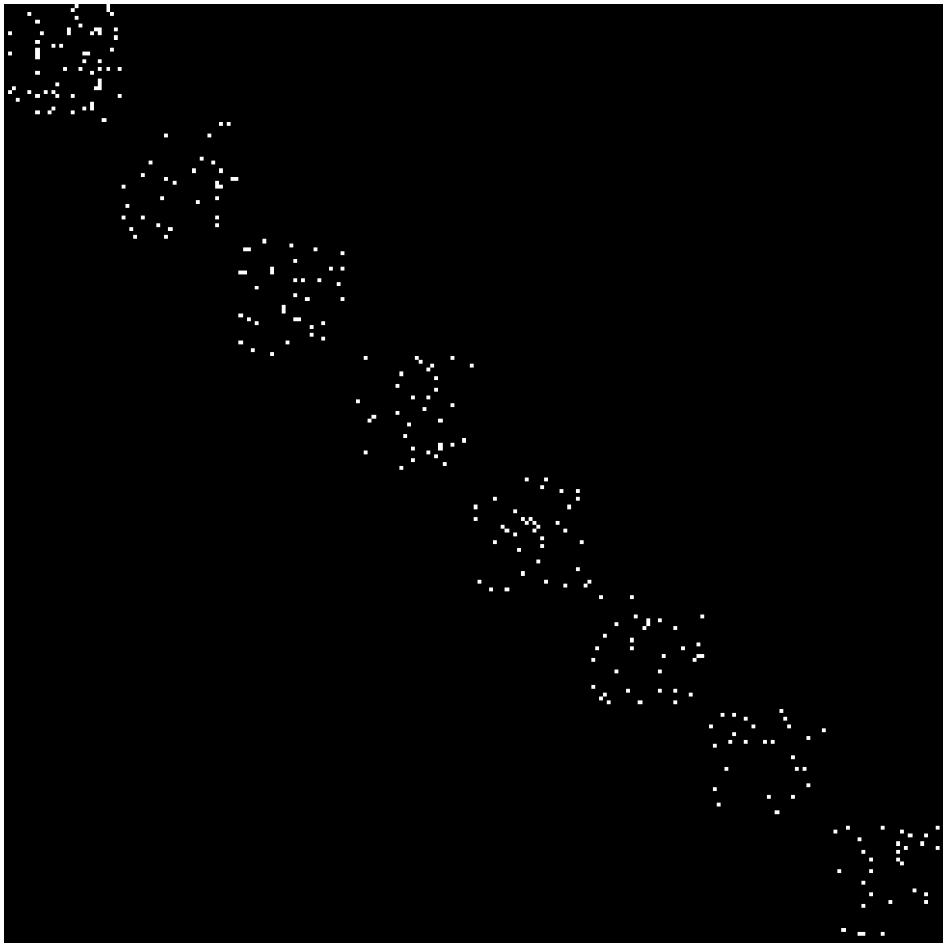}\hspace{1cm}\includegraphics[width=0.3\linewidth]{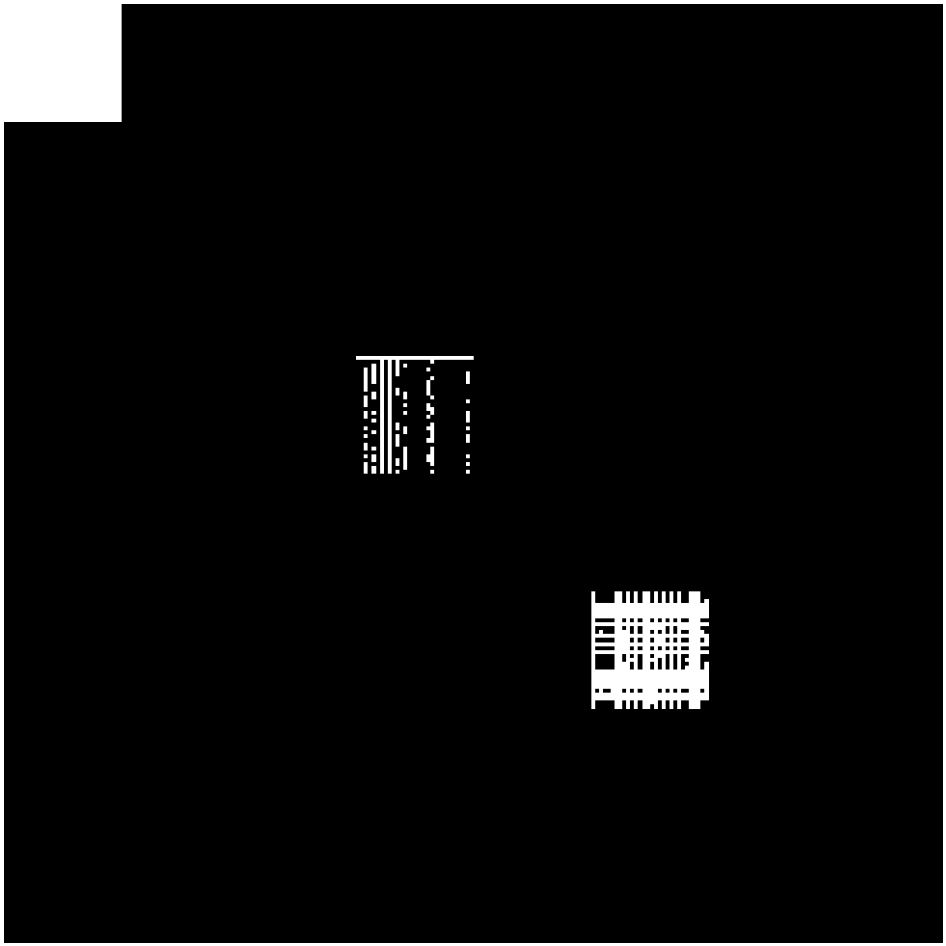}\\
(unstructured pruning) \hspace{1.2cm} (semi-structured pruning) 
\caption{This figure shows a crop of the mask tensor obtained after the gating parametrization when trained on the FPHA dataset. Top-left corresponds to the original mask (without pruning) while the others  correspond to masks obtained with structured,  unstructured  and semi-structured pruning respectively.   In all these masks,  each diagonal block corresponds to a channel.  Better to zoom the PDF.}\label{fig:A3}
\end{figure*}

\section{Experiments}\label{section5}

This section assesses baseline and pruned GCNs' performance in skeleton-based recognition using SBU Interaction~\cite{Yun2012} and the First Person Hand Action (FPHA)~\cite{Garcia2018} datasets, comparing our lightweight GCNs against related pruning techniques. SBU is an interaction dataset acquired using the Microsoft Kinect sensor, it contains 282 moving skeleton sequences performed by two interacting individuals and belonging to 8 categories. Each pair of interacting individuals corresponds to two 15 joint skeletons, each one encoded with a sequence of its 3D coordinates across video frames. The evaluation protocol follows the train-test split as in the original dataset release~\cite{Yun2012}. The FPHA dataset includes 1175 skeletons belonging to 45 action categories performed by 6 different individuals in 3 scenarios. Action categories are highly variable, including various styles, speed, scale, and viewpoint. Each skeleton includes 21 hand joints, each one again encoded with a sequence of its 3D coordinates across video frames. The performances of different methods are evaluated using the 1:1 setting proposed in~\cite{Garcia2018} with 600 action sequences for training and 575 for testing. The average accuracy over all classes of actions is reported in all experiments. \\ 
\begin{table}
  \begin{minipage}[c]{0.45\textwidth}
\centering
\vspace{1.91cm}
\resizebox{0.86\columnwidth}{!}
{
\begin{tabular}{cc|c}
{\bf Method}      &   & {\bf Accuracy (\%)}\\
\hline 
  Raw Position \cite{Yun2012} & $ \ $   & 49.7   \\ 
  Joint feature \cite{Ji2014}  & $ \ $   & 86.9   \\
  CHARM \cite{Li2015a}       & $ \ $    & 86.9   \\
 \hline  
H-RNN \cite{Du2015}         & $ \ $    & 80.4   \\ 
ST-LSTM \cite{Liu2016}      & $ \ $    & 88.6    \\ 
Co-occurrence-LSTM \cite{Zhua2016} & $ \ $  & 90.4  \\ 
STA-LSTM  \cite{Song2017}     & $ \ $   & 91.5  \\ 
ST-LSTM + Trust Gate \cite{Liu2016} & $ \ $  & 93.3 \\
VA-LSTM \cite{Zhang2017}      & $ \ $  & 97.6  \\
 GCA-LSTM \cite{GCALSTM}                    &   $ \ $      &  94.9     \\ 
  \hline
Riemannian manifold. traj~\cite{RiemannianManifoldTraject} &  $ \ $  & 93.7 \\
DeepGRU  \cite{DeepGRU}        &    $ \ $   &    95.7    \\
RHCN + ACSC + STUFE \cite{Jiang2020} & $ \ $   & \bf98.7 \\ 
  \hline
\hline 
  Our baseline GCN &              &        \underline{98.4}      
\end{tabular}}
\vspace{0.25cm}
\captionof{table}{Comparison of our baseline GCN against related work on the SBU database.  Results shown in Bold stand for the best performances while those underlined correspond to the second best performances. }\label{tab222}
\end{minipage}
\begin{minipage}[c]{0.45\textwidth}
\centering
\resizebox{1\columnwidth}{!}{
\begin{tabular}{ccccc}
{\bf Method} & {\bf Color} & {\bf Depth} & {\bf Pose} & { \bf Accuracy (\%)}\\
\hline
  2-stream-color \cite{refref10}   & \cmark  &  \xmark  & \xmark  &  61.56 \\
 2-stream-flow \cite{refref10}     & \cmark  &  \xmark  & \xmark  &  69.91 \\  
 2-stream-all \cite{refref10}      & \cmark  & \xmark   & \xmark  &  75.30 \\
\hline 
HOG2-dep \cite{refref39}        & \xmark  & \cmark   & \xmark  &  59.83 \\    
HOG2-dep+pose \cite{refref39}   & \xmark  & \cmark   & \cmark  &  66.78 \\ 
HON4D \cite{refref40}               & \xmark  & \cmark   & \xmark  &  70.61 \\ 
Novel View \cite{refref41}          & \xmark  & \cmark   & \xmark  &  69.21  \\ 
\hline
1-layer LSTM \cite{Zhua2016}        & \xmark  & \xmark   & \cmark  &  78.73 \\
2-layer LSTM \cite{Zhua2016}        & \xmark  & \xmark   & \cmark  &  80.14 \\ 
\hline 
Moving Pose \cite{refref59}         & \xmark  & \xmark   & \cmark  &  56.34 \\ 
Lie Group \cite{Vemulapalli2014}    & \xmark  & \xmark   & \cmark  &  82.69 \\ 
HBRNN \cite{Du2015}                & \xmark  & \xmark   & \cmark  &  77.40 \\ 
Gram Matrix \cite{refref61}         & \xmark  & \xmark   & \cmark  &  85.39 \\ 
TF    \cite{refref11}               & \xmark  & \xmark   & \cmark  &  80.69 \\  
\hline 
JOULE-color \cite{refref18}         & \cmark  & \xmark   & \xmark  &  66.78 \\ 
JOULE-depth \cite{refref18}         & \xmark  & \cmark   & \xmark  &  60.17 \\ 
JOULE-pose \cite{refref18}         & \xmark  & \xmark   & \cmark  &  74.60 \\ 
JOULE-all \cite{refref18}           & \cmark  & \cmark   & \cmark  &  78.78 \\ 
\hline 
Huang et al. \cite{Huangcc2017}     & \xmark  & \xmark   & \cmark  &  84.35 \\ 
Huang et al. \cite{ref23}           & \xmark  & \xmark   & \cmark  &  77.57 \\  
\hline 
HAN  \cite{Liu2021}   & \xmark  & \xmark   & \cmark & \underline{85.74} \\
  \hline
  \hline
Our  baseline GCN                   & \xmark  & \xmark   & \cmark  &  \bf86.43                                                 
\end{tabular}}
\vspace{0.25cm}
\captionof{table}{Comparison of our baseline GCN against related work on the FPHA database.}\label{compare2}
\end{minipage}
\end{table}

 \begin{table}[h]
 \begin{center}
\resizebox{0.59\columnwidth}{!}{
  \begin{tabular}{clll}    
   \rotatebox{0}{Pruning rates}  &     \rotatebox{0}{Accuracy (\%)} & SpeedUp  & \rotatebox{0}{Observation}  \\
 \hline
  \hline
    0\%    &    \bf98.40   & none  & Baseline GCN\\
 
                                70\% &  93.84 & none  &  Band-stop Weight Param.\\

    \hline
    \multirow{6}{*}{\rotatebox{0}{90\%}}     & 87.69  &  \bf426$\times$ & Structured  \\
                                             & 89.23  &  \bf487$\times$ & Structured  (+ rank optimization)   \\
                                                                  &   \bf93.84  & none   & Unstructured    \\
                               &   \bf93.84  &16$\times$   & Unstructured  (+ rank optimization)     \\
                                 &   \underline{90.76}  &  \underline{40}$\times$ & Semi-structured             \\
                                 &   \underline{89.23}  &  \underline{52}$\times$ & Semi-structured   (+ rank optimization)             \\
    \hline
    \multirow{6}{*}{\rotatebox{0}{95\%}}     & 87.69  &  \bf678$\times$ & Structured  \\
                                              & 87.69  &  \bf787$\times$ & Structured  (+ rank optimization)   \\
                                                                  &   \bf92.30   & none    & Unstructured    \\
                                &   \underline{92.30}   & 16$\times$   & Unstructured   (+ rank optimization)    \\
                                 &   \bf92.30  &  \underline{109}$\times$ & Semi-structured             \\
                                 &   \bf93.84  &  \underline{106}$\times$ & Semi-structured   (+ rank optimization)             \\
\hline 
    \multirow{6}{*}{\rotatebox{0}{98\%}}     &  81.53   & \bf797$\times$ & Structured  \\
                                             &  81.53   & \bf2195$\times$ & Structured  (+ rank optimization)   \\
                                                                  &   \bf89.23 &none  &  Unstructured    \\
                               &   \bf89.23 & 106$\times$  &  Unstructured  (+ rank optimization)     \\
                                 &   \underline{83.07}   &  \underline{135}$\times$ & Semi-structured             \\
                                 &   \underline{86.15}   &  \underline{607}$\times$ & Semi-structured  (+ rank optimization)              \\
    \hline \hline
  \multicolumn{4}{c}{Comparative (regularization-based) pruning}   \\                              
    \hline 
      \multirow{4}{*}{\rotatebox{0}{98\%}}                                &    55.38 & none & MP+$\ell_0$-reg. \\
                                                         &    73.84 & none  & MP+$\ell_1$-reg. \\                                                                                                                                                                      
                                 &    61.53 & none &  MP+Entropy-reg. \\ 
                                &   75.38 & none & MP+Cost-aware-reg.

  \end{tabular}}
\end{center}
\caption{This table shows detailed performances and ablation study on SBU for different  pruning rates. ``none'' stands for no-actual speedup is observed as the underlying tensors/architecture remain  shaped identically to the unpruned network (despite having pruned connections).  For structured,  unstructured and semi-structured settings,   when ``rank optimization'' is not used,  only pruning rate is considered in the loss together with cross entropy.  When ``rank optimization'' is used,  all the three terms are combined in the loss. }\label{table21}
\end{table}
 \begin{table}[h]
 \begin{center}
\resizebox{0.59\columnwidth}{!}{
  \begin{tabular}{clll}    
   \rotatebox{0}{Pruning rates}  &     \rotatebox{0}{Accuracy (\%)} & SpeedUp  & \rotatebox{0}{Observation}  \\
 \hline
  \hline
    0\%    &    \bf86.43   & none  & Baseline GCN\\
 
                                50\% & 85.56 &  none & Band-stop Weight Param.\\

    \hline
    \multirow{6}{*}{\rotatebox{0}{90\%}}     & 68.00  &  \bf274$\times$ & Structured  \\
                                              & 71.30  &  \bf547$\times$ & Structured (+ rank optimization) \\
                                                                  &  \bf83.82  & none  & Unstructured    \\
                                 &  \bf84.17   & 16$\times$  & Unstructured  (+ rank optimization)   \\
                                 &   \underline{78.60}   & \underline{33}$\times$ & Semi-structured             \\
                                 &   \underline{80.52}   & \underline{38}$\times$ & Semi-structured  (+ rank optimization)            \\
    \hline
    \multirow{6}{*}{\rotatebox{0}{95\%}}     & 56.69  &   \bf759$\times$ & Structured  \\
                                             & 62.60  &   \bf931$\times$ & Structured (+ rank optimization)  \\
                                                                  &    \bf78.78   & none  & Unstructured    \\
                                 &    \bf80.17   & 29$\times$  & Unstructured (+ rank optimization)    \\
                                 &   \underline{72.17}   &  \underline{197}$\times$ & Semi-structured             \\
                                 &   \underline{74.60}   &  \underline{214}$\times$ & Semi-structured  (+ rank optimization)           \\ 
\hline 
    \multirow{6}{*}{\rotatebox{0}{98\%}}     & 47.47  &   \bf1479$\times$ & Structured  \\
                      & 49.04  &   \bf1399$\times$ & Structured  (+ rank optimization) \\
                                                                  &    \bf78.08   & none   & Unstructured    \\
                                 &    \bf77.56   & 126$\times$  & Unstructured (+ rank optimization)    \\
                                 &   \underline{75.13}   &  \underline{33}$\times$ & Semi-structured             \\
                                 &   \underline{73.91}   &  \underline{278}$\times$ & Semi-structured  (+ rank optimization)            \\
    \hline \hline
  \multicolumn{4}{c}{Comparative (regularization-based) pruning}   \\                              
    \hline 
      \multirow{4}{*}{\rotatebox{0}{98\%}}                                      &    64.69 & none  & MP+$\ell_0$-reg. \\
                                                         &   70.78 & none & MP+$\ell_1$-reg. \\                                                                                                                                                                      
                                 &    67.47 & none & MP+Entropy-reg. \\ 
                                &   69.91 & none & MP+Cost-aware-reg.

  \end{tabular}}
\end{center}
\caption{This table shows detailed performances and ablation study on FPHA for different  pruning rates. ``none'' stands for no-actual speedup is observed as the underlying tensors/architecture remain shaped identically to the unpruned network (despite having pruned connections).}\label{table22}
\end{table}

\noindent {\bf Input graphs.}  Let's consider a sequence of skeletons $\{S^t\}_t$ with $S^t=\{\hat{p}_1^t,\dots, \hat{p}_n^t\}$ being the 3D skeleton coordinates at frame $t$, and $\{\hat{p}_j^t\}_t$ a joint trajectory  through successive frames. We define an input graph $\G = (\V,\E)$ as a finite collection of trajectories, with each node $v_j \in \V$ in $\G$ being a trajectory $\{\hat{p}_j^t\}_t$, and an edge  $(v_j, v_i) \in  \E$  exists between two nodes if the underlying trajectories are spatially neighbors. Each trajectory is processed using {\it temporal chunking}, which splits the total duration of a sequence into $M$ evenly-sized temporal chunks (with $M=4$ in practice). Then, joint coordinates  $\{\hat{p}_j^t\}_t$ of the trajectory are assigned to these chunks, based on their time stamps. The averages of these chunks are concatenated in order to create the raw description of the trajectory (denoted as $\phi(v_j) \in \mathbb{R}^{s}$ with $s=3 \times M$). This process preserves the temporal structure of trajectories while being frame-rate and duration agnostic.\\

\noindent {\bf Implementation details \& baseline GCNs.}  All the GCNs have been trained using the Adam optimizer for $2,700$ epochs with a batch size of $200$ for SBU and $600$ for FPHA, a momentum of $0.9$, and a global learning rate  (denoted as $\nu(t)$)  inversely proportional to the speed of change of the loss used to train the networks; with $\nu(t)$ decreasing as $\nu(t) \leftarrow \nu(t-1) \times 0.99$ (resp. increasing as $\nu(t) \leftarrow \nu(t-1) \slash 0.99$) when the speed of change of the loss in Eq.~\ref{eq4} increases (resp. decreases). Experiments were run on a GeForce GTX 1070 GPU device with 8 GB memory, without dropout or data augmentation. The baseline GCN architecture for SBU includes an attention layer of 8 heads, a convolutional layer of 16 filters, a dense fully connected layer, and a softmax layer. The baseline GCN architecture for FPHA is heavier and includes 16 heads, a convolutional layer of 32 filters, a dense fully connected layer, and a softmax layer. Both the baseline GCN architectures, on the SBU and the FPHA benchmarks, are accurate  (see tables.~\ref{tab222} and~\ref{compare2}), and our goal is to make them lightweight while maintaining their high accuracy.\\

\noindent{\bf Performances, Comparison \& Ablation.} Tables~\ref{table21}-\ref{table22}  show a comparison and an ablation study of our method both on the SBU and the FPHA datasets.  First,  according to the observed results,  when only the cross entropy loss is used without budget   (i.e., $\lambda=\beta=0$ in  Eq. \ref{eq4}), performances are close  to the initial heavy GCNs (particularly on FPHA), with less parameters\footnote{Pruning rate does not exceed 70\% and no control on this rate is achievable when $\lambda=0$.} as this produces a regularization effect similar to \cite{dropconnect2013}. Then,  when pruning is structured, the accuracy is relatively low but the speedup is important particularly for high pruning regimes.  When pruning is unstructured, the accuracy reaches its highest value,  but no actual speedup is observed as the architecture of the pruned networks remains unchanged (i.e., not compact).   When pruning is semi-structured,  we observe the best trade-off between accuracy and speedup;  in other words, {\it coarsely} pruned parts of the network (related to entire block/column/row connections) lead to high speedup and efficient computation, whereas {\it finely} pruned parts (related to individual connections) lead to a better accuracy with a contained marginal impact on computation, so speedup is still globally observed with a significant amount.  \\
\indent Extra comparison of our method against other regularizers shows a substantial gain. Indeed, our method is compared against different variational pruning with regularizers  plugged in Eq. \ref{eq4} (instead of our proposed budget and rank regularizers),  namely $\ell_0$ \cite{REFICLR},  $\ell_1$ \cite{refref74},  entropy \cite{refref75} and $\ell_2$-based cost  \cite{REFLemaire}, all without our parametrization. From the observed results, the impact of our method  is substantial for different settings  and for  equivalent pruning rate (namely 98\%).  Note that when alternative regularizers are used, multiple settings (trials) of the underlying mixing hyperparameters (in Eq. \ref{eq4}) are considered prior to reach the targeted pruning rate, and this makes the whole training and pruning process overwhelming.  While cost-aware regularization makes training more affordable, its downside resides in the observed collapse of trained masks; this is a well known effect that affects performances  at high pruning rates.   Finally,  Fig.  \ref{fig:A3} shows examples of obtained mask tensors taken from the second (attention) layer of the pruned GCN.  For semi-structured pruning,  we observe a compact tensor layer with some individually pruned connections whereas structured and unstructured pruning --- when applied separately --- either produce {\it compact} or {\it spread} tensors, with a negative impact on respectively {\it accuracy} or {\it speed}. In sum, semi-structured pruning gathers the advantages of {\it both} while discarding their inconveniences.

\section{Conclusion}
This paper introduces a novel magnitude pruning approach that combines {\it both} the strengths of structured and unstructured pruning methods while discarding their drawbacks. The proposed method, dubbed as  {\it semi-structured}, is based on a novel cascaded weight parametrization including band-stop, weight-sharing, and gating mechanisms. Our pruning method also relies on a budget loss that allows implementing  fine-grained targeted pruning rates while also reducing the rank of the pruned tensors resulting in more efficient and still effective networks. Extensive experiments, conducted on the challenging task of skeleton-based recognition, corroborate all these findings.

\end{document}